\pdfoutput=1

\documentclass[11pt]{article}

\usepackage[final]{acl}

\usepackage{times}
\usepackage{latexsym}

\usepackage[T1]{fontenc}

\usepackage[utf8]{inputenc}

\usepackage{microtype}

\usepackage{inconsolata}

\usepackage{graphicx}

\usepackage{booktabs}
\usepackage{subcaption}
\usepackage{multirow}
\usepackage{amsmath}
\usepackage{pifont}
\usepackage{float}

\title{Corrective In-Context Learning:\\Evaluating Self-Correction in Large Language Models}

\author{
    Mario Sanz-Guerrero$^1$ \and Katharina von der Wense$^{1,2}$ \\
    $^1$Johannes Gutenberg University Mainz, Germany \\
    $^2$University of Colorado Boulder, USA \\
    \texttt{\{\href{mailto:msanzgue@uni-mainz.de}{msanzgue}, \href{mailto:k.vonderwense@uni-mainz.de}{k.vonderwense}\}@uni-mainz.de}
}

\begin{document}
\maketitle
\begin{abstract}
In-context learning (ICL) has transformed the use of large language models (LLMs) for NLP tasks, enabling few-shot learning by conditioning on labeled examples without finetuning. Despite its effectiveness, ICL is prone to errors, especially for challenging examples. With the goal of improving the performance of ICL, we propose \textit{corrective in-context learning} (CICL), an approach that incorporates a model's incorrect predictions alongside ground truth corrections into the prompt, aiming to enhance classification accuracy through self-correction. However, contrary to our hypothesis, extensive experiments on text classification tasks demonstrate that CICL consistently underperforms standard ICL, with performance degrading as the proportion of corrections in the prompt increases. Our findings indicate that CICL introduces confusion by disrupting the model's task understanding, rather than refining its predictions. Additionally, we observe that presenting harder examples in standard ICL does not improve performance, suggesting that example difficulty alone may not be a reliable criterion for effective selection. By presenting these negative results, we provide important insights into the limitations of self-corrective mechanisms in LLMs and offer directions for future research.\footnote{Code and data are available at \url{https://github.com/mario-sanz/CICL}.}

\end{abstract}

\section{Introduction}

In-context learning \citep[ICL;][]{brown2020fewshot} has emerged as a powerful paradigm for leveraging large language models (LLMs) for various NLP tasks, including text classification. Unlike traditional approaches that require finetuning on task-specific data, ICL allows models to make predictions based on a small number of examples presented in the prompt, effectively transforming LLMs into flexible tools for few-shot learning. This paradigm has demonstrated remarkable performance in numerous scenarios, often approaching or surpassing finetuned models on specific tasks \citep{brown2020fewshot}.

Although effective, ICL is susceptible to errors, particularly with difficult examples. We set out to further improve ICL by introducing a novel extension of it, which we term \textit{corrective in-context learning} (CICL). Our approach is based on the idea that providing the model with its initial predictions alongside the correct ground truth labels can serve as a feedback mechanism, enabling the model to refine its understanding and improve subsequent predictions.

\begin{figure}
    \centering
    \includegraphics[width=0.89\linewidth]{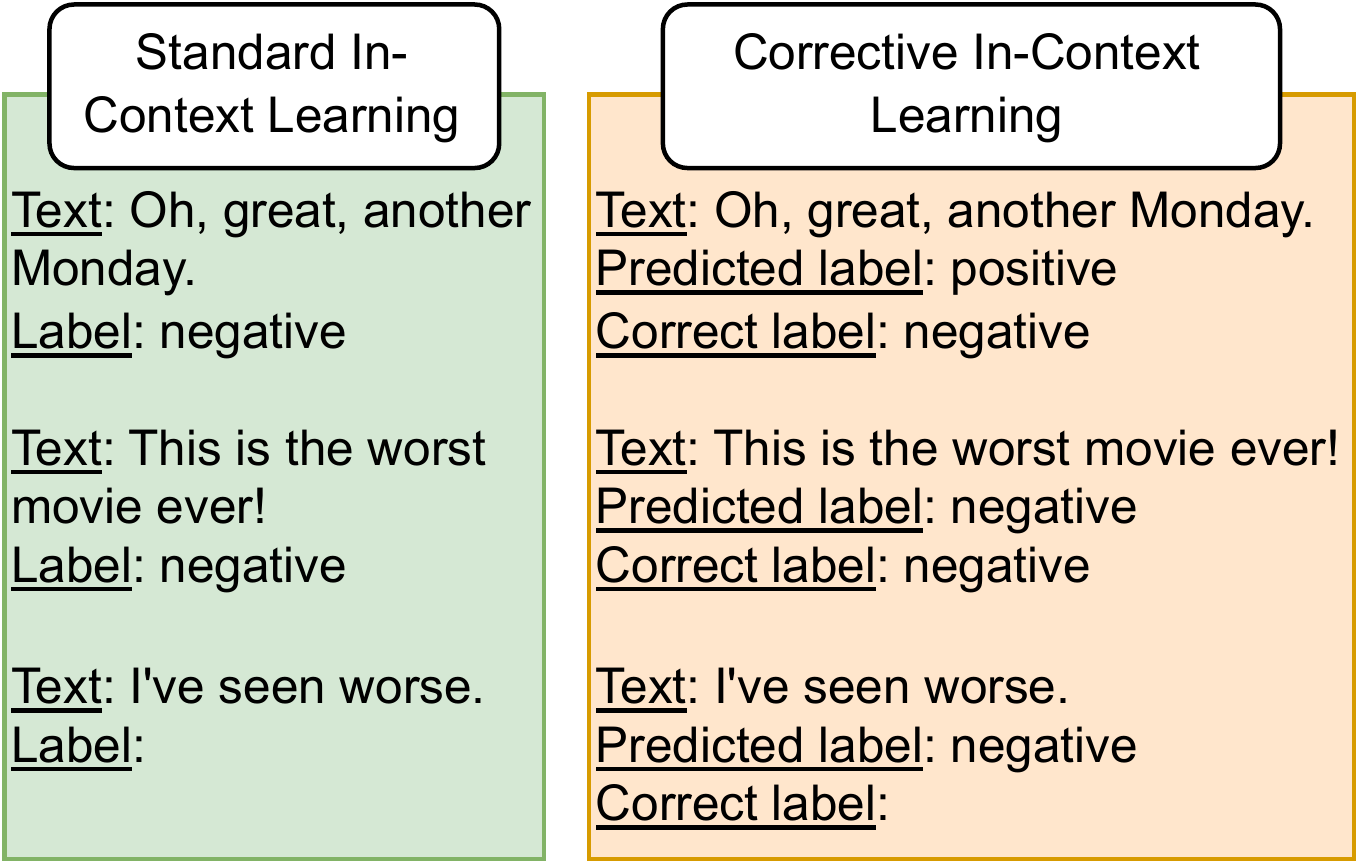}
    \caption{CICL prompt example. The model is tasked with predicting the correct label based on its own prediction, using examples that include both the prediction and its correction.}
    \label{fig:enter-label}
\end{figure}

CICL builds on the intuition that LLMs, when exposed to their own errors in conjunction with the correct answers, might learn from these mistakes within the confines of a single interaction. For instance, if a model predicts ``positive'' for a sentence that is actually ``negative,'' presenting this prediction--error pair could prompt the model to recalibrate its internal representations and make better-informed predictions for similar inputs.

However, our empirical evaluation reveals a different story. While the approach is theoretically promising, our experiments show that CICL fails to deliver the anticipated improvements. In some cases, performance even deteriorates compared to standard ICL. This underscores the complexity of self-correction in LLMs and the importance of rigorously evaluating intuitive extensions to ICL.

In this paper, we present a detailed investigation of CICL for text classification tasks. We outline our methodology, describe the experimental setup, and analyze the results, focusing on understanding why this approach falls short of expectations. By sharing these negative results, we aim to contribute to the growing discourse on the limitations of LLMs and to inspire future research to develop more effective self-correction mechanisms.

\section{Related Work}
\label{app:related_work}

\paragraph{In-Context Learning}
Introduced by \citet{brown2020fewshot}, ICL enables LLMs to perform tasks via few-shot prompting without parameter updates. Subsequent work has explored factors influencing ICL performance, including example selection \citep{liu2022gpt3examples} and ordering \citep{lu2022fantastically}, raising questions about its robustness to errors or ambiguous examples. To enhance ICL, prompt tuning methods \citep{lester2021power,li2021prefix} optimize soft prompts, improving adaptation while maintaining frozen model parameters, bridging the gap between ICL and finetuning.

\paragraph{Self-Correction in LLMs}
Prior work explores self-correction through iterative refinement \citep{madaan2023selfrefine}, finetuning on self-generated data \citep{huang2023selfimprove}, or reinforcement learning \citep{kumar2024rl}. These approaches, however, require multi-step processes, parameter updates, or external rewards. In contrast, we investigate whether LLMs can self-correct \emph{in-context} by directly incorporating corrections into the prompt. Closest to our setting, \citet{monea2024icrl} show that LLMs struggle to improve from binary reward signals (correct/incorrect) in ICL scenarios. While their feedback is implicit, we extend this observation to \emph{explicit ground truth corrections} and similarly find degraded performance.

\section{Methodology}

\begin{figure*}
    \centering
    \includegraphics[width=\linewidth]{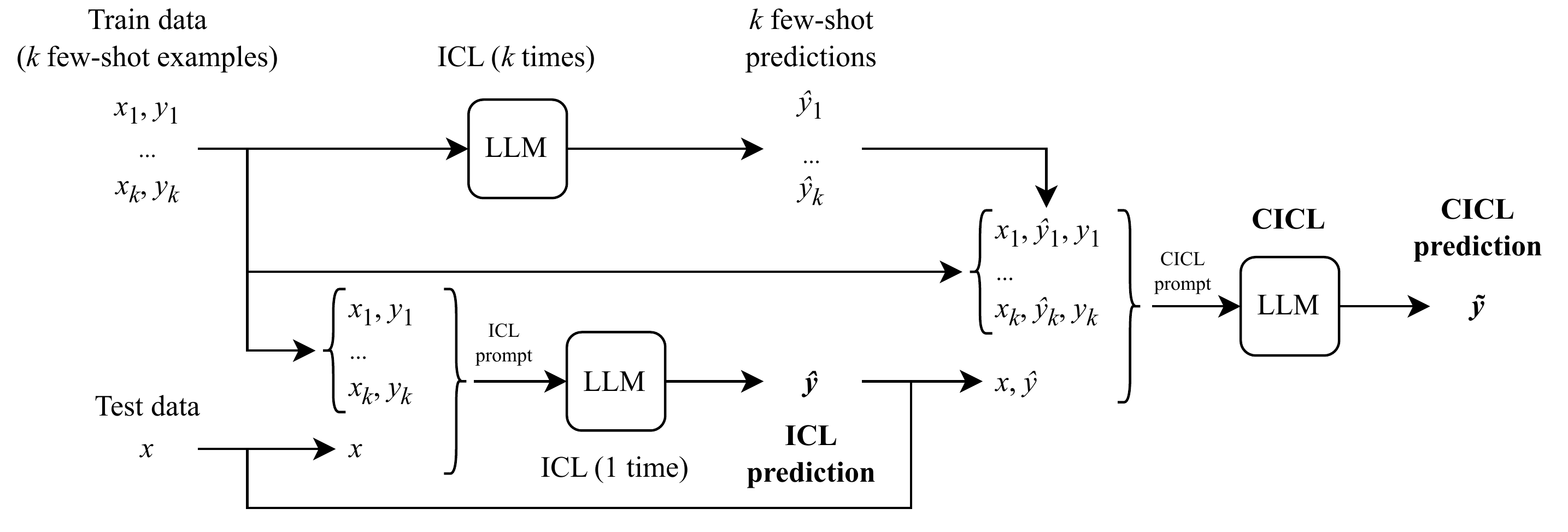}
    \caption{Diagram of the CICL methodology. For each test instance $x$, $k$ few-shot examples $(x_1, y_1), \dots, (x_k, y_k)$ are selected. Standard ICL generates predictions $\hat{y}_1, \dots, \hat{y}_k$ for these examples, which are used to build the CICL prompt. This, combined with the initial ICL prediction $\hat{y}$ for $x$, allows the model to predict the corrected label $\tilde{y}$.}
    \label{fig:methodology}
\end{figure*}

\subsection{Standard In-Context Learning}

ICL leverages the few-shot capabilities of LLMs by conditioning predictions on a prompt constructed from a small set of labeled examples. Formally, given a dataset of examples $\{x_i, y_i\}$, where $x_i$ represents a text input and $y_i$ is a class label verbalized as one of the possible labels in the set $\mathcal{L}$, the model $M$ predicts the label for a query $x$ using a context prompt $C_k$ of $k$ examples:

\begin{equation}\label{eq:pred}
    \hat{y} = \mathop{\arg \max}\limits_{y \in \mathcal{L}} P_M(y \mid C_k, x)
\end{equation}

ICL operates under the assumption that the few-shot examples encapsulate task-relevant patterns, allowing the model to generalize to unseen queries. This paradigm has demonstrated remarkable versatility and competitive performance across numerous tasks and datasets \citep{brown2020fewshot}. However, its effectiveness is heavily dependent on the quality and representativeness of the examples in the prompt \citep{liu2022gpt3examples}. Furthermore, biases in the selection or ordering of examples, as well as the inherent biases of the model, can significantly influence predictions, posing challenges for consistent and reliable performance \citep{zhao2021calibrate}.

In this study, standard ICL is the baseline against which we compare the proposed CICL approach, providing insights into whether iterative feedback mechanisms can address some of these challenges. The structure of the ICL prompt, along with an example, is provided in Appendix \ref{app:icl_prompt}.

\subsection{Corrective In-Context Learning}

CICL extends the standard ICL paradigm by introducing a second round of predictions informed by feedback from the first round. The algorithm is as follows:

\begin{enumerate}
    \item \textit{Initial prediction}:
    A prompt is constructed from $k$ randomly selected examples from the training data, each consisting of an input--output pair $(x_i, y_i)$, for $i \in {1, ..., k}$. This prompt is used to generate prediction $\hat{y}$ for the test input $x$ via standard ICL.
    
    \item \textit{Feedback incorporation}:
    We perform ICL for the $k$ examples selected in Step 1, getting the model's predictions $\hat{y}_i$ for each example $(x_i, y_i)$, with the remaining $k-1$ examples acting as few-shot examples. Using these predictions, a feedback-augmented prompt (CICL) is constructed from triplets of the form $(x_i, \hat{y}_i, y_i)$, where $x_i$ is the input text, $\hat{y}_i$ is the predicted label, and $y_i$ is the true label. These triplets explicitly highlight the model's errors (when $\hat{y}_i \neq y_i$) and correct predictions (when $\hat{y}_i = y_i$), providing the model with a context to learn from its earlier outputs.
    
    \item \textit{Corrective prediction}:
    To perform the CICL prediction for the input $x$, the feedback prompt built in Step 2 is expanded to include $x$ along with its initial prediction $\hat{y}$ (obtained in Step 1). The task for the model is then to predict a corrected label $\tilde{y}$ for $x$, leveraging the feedback triplets to generalize how errors were corrected in the few-shot examples.
\end{enumerate}

This iterative setup allows the model to ``see'' its own mistakes and explicitly incorporate the correct answers during the second round, hypothesizing that this feedback mechanism can improve its performance. Figure \ref{fig:methodology} shows a diagram of the proposed methodology for CICL, and the structure of the CICL prompt, along with an example, is provided in Appendix \ref{app:cicl_prompt}.

\section{Experiments}

\subsection{Experimental Setup}
\paragraph{Datasets}
We evaluate CICL using 17 text classification datasets widely adopted in previous work. These datasets span a variety of tasks, including sentiment analysis, topic classification, and more. Further details are provided in Appendix \ref{app:datasets}.

\paragraph{Models}
To explore the effectiveness of CICL across different models and ensure findings are not model-specific, we use four recent LLMs: Llama-3.1 \citep[8B;][]{dubey2024llama3}, GPT-J \citep[6B;][]{wang2021gptj}, Mistral 7B v0.3 \citep{jiang2023mistral7b} and Qwen2.5 \citep[7B;][]{qwen2024qwen25}. The choice of these relatively small-sized models allows for extensive experimentation while maintaining computational feasibility. However, preliminary experiments with the larger 70B version of Llama-3.1 yielded similar results to the smaller versions.

\paragraph{Implementation Details}
Following prior work on ICL for text classification, we use simple and unified templates for all datasets and do not include task instructions, keeping human engineering to a minimal level \citep{min2022rethinking, fei2023domaincalibration}. Also following prior work, we set $k=8$ few-shot examples, which enables incorporating a fair number of corrections in the prompt while keeping computational costs manageable. Preliminary experiments with larger $k$ values showed similar results, so we stick with $k=8$ for simplicity.

To assess how CICL performs with different levels of corrected examples, we introduce varying proportions of corrected examples in the CICL prompt, ranging from 0\% (no corrected examples) to 100\% (all examples corrected) in increments of 25\%. Each proportion determines how many of the $k$ examples in the CICL prompt are corrected (i.e., their initial ICL prediction was incorrect). This approach allows us to evaluate how the ratio of corrected feedback influences the model's ability to refine its predictions. To minimize the impact of randomness in the results, every experimental configuration is run using 5 different random seeds.

\paragraph{Metric}
We compare standard ICL and CICL performance using macro-F1 score, which accounts for class imbalance.

\subsection{Results}

\begin{figure*}
    \centering
    \begin{subfigure}[b]{0.24\textwidth}
        \centering
        \includegraphics[width=\linewidth]{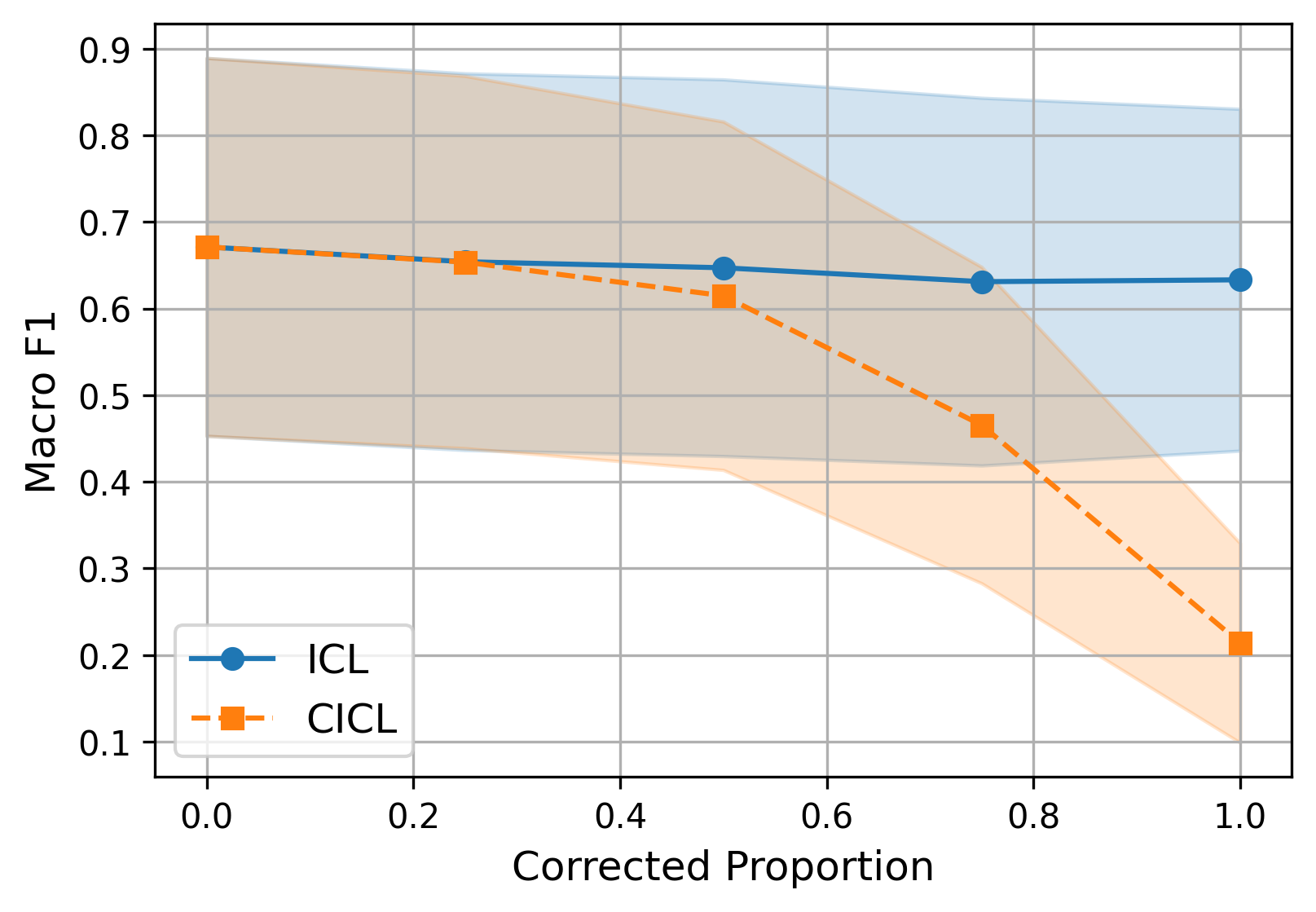}
        \caption{Llama-3.1 (8B).}
        \label{fig:sub1}
    \end{subfigure}
    \hfill
    \begin{subfigure}[b]{0.24\textwidth}
        \centering
        \includegraphics[width=\linewidth]{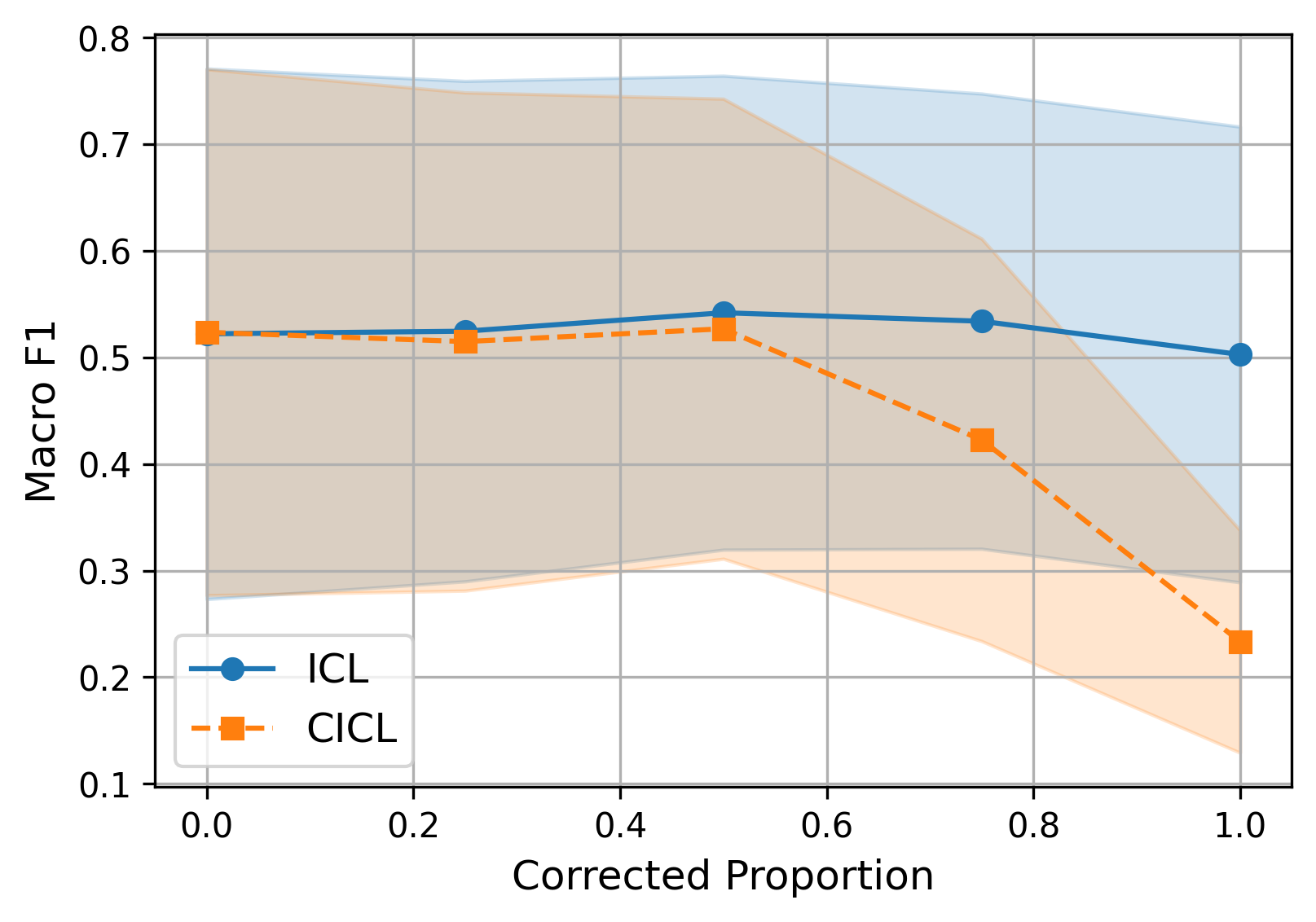}
        \caption{GPT-J (6B).}
        \label{fig:sub2}
    \end{subfigure}
    \hfill
    \begin{subfigure}[b]{0.24\textwidth}
        \centering
        \includegraphics[width=\linewidth]{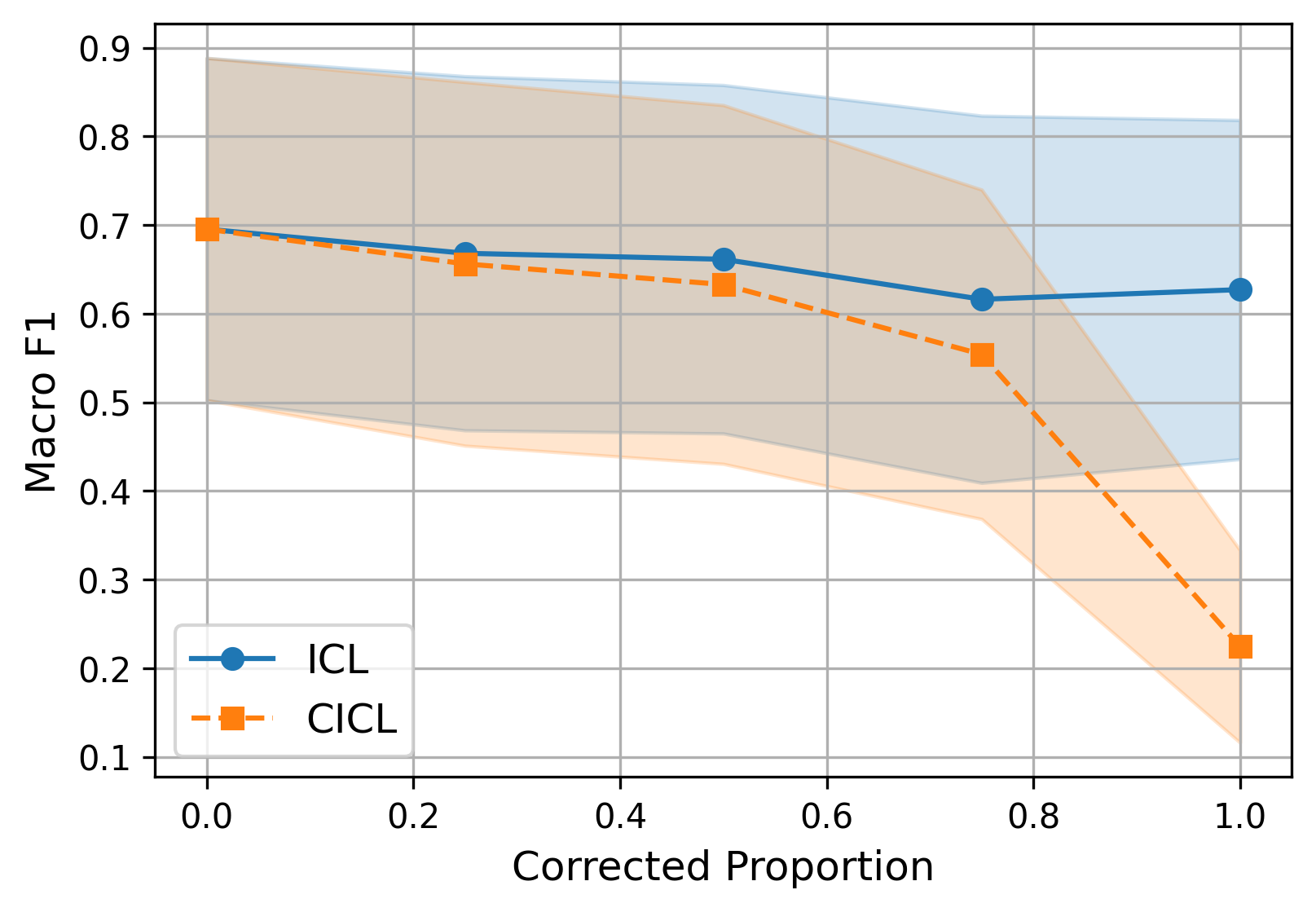}
        \caption{Mistral 7B v0.3.}
        \label{fig:sub3}
    \end{subfigure}
    \hfill
    \begin{subfigure}[b]{0.24\textwidth}
        \centering
        \includegraphics[width=\linewidth]{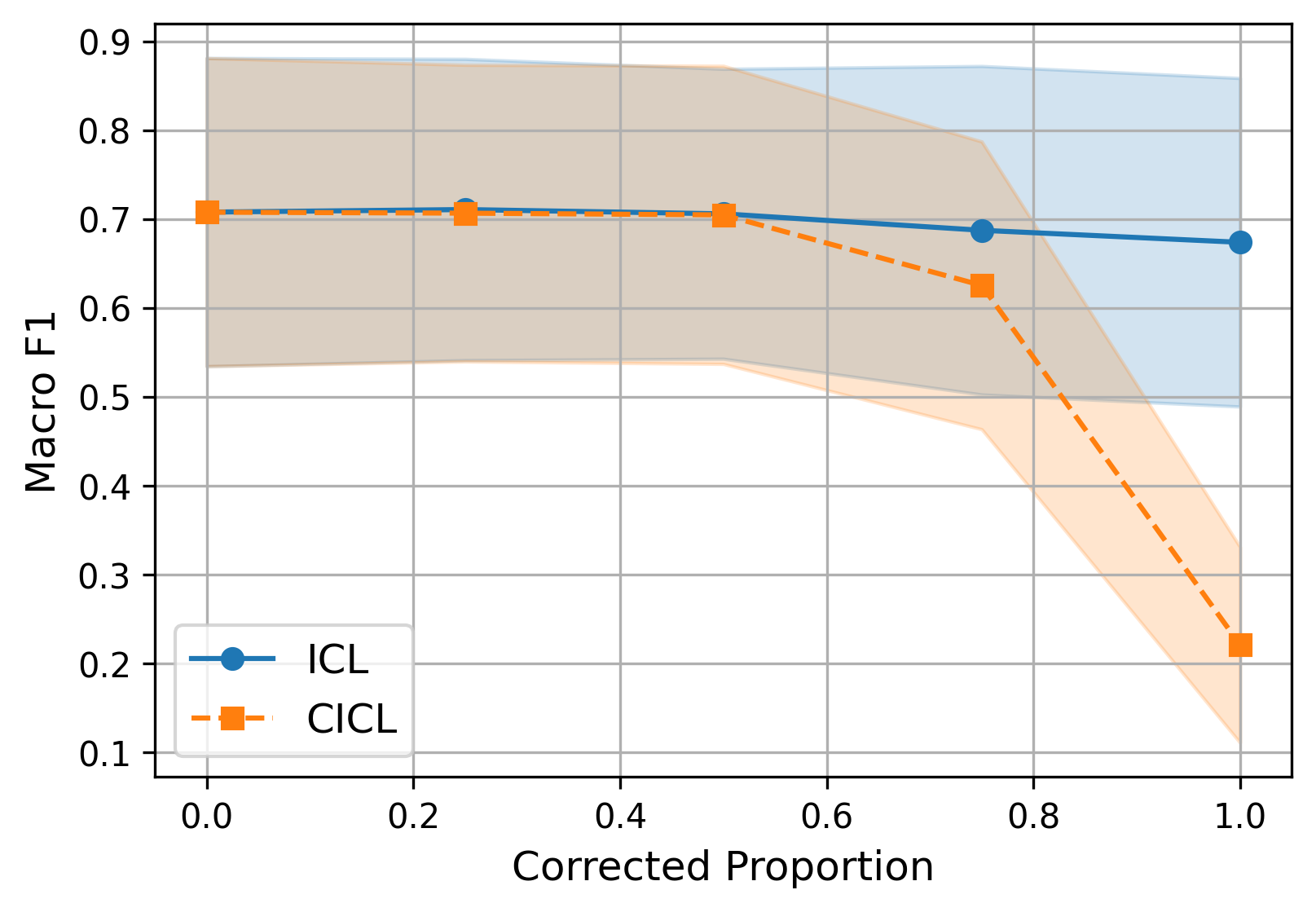}
        \caption{Qwen2.5 (7B).}
        \label{fig:sub4}
    \end{subfigure}
    \caption{Mean Macro-F1 ($\pm$ Std Dev) across all datasets for each model, comparing standard ICL (blue) and CICL (orange). As the proportion of corrected examples increases ($x$-axis), the performance of CICL decreases ($y$-axis).}
    \label{fig:results}
\end{figure*}

Figure \ref{fig:results} compares the performance of standard ICL and CICL across different models. The results present the mean and standard deviation of Macro-F1 scores across all datasets, evaluated for varying proportions of corrected examples in CICL. Detailed results are available in Appendix \ref{app:detailed_results}.

Contrary to our hypothesis, CICL consistently underperforms standard ICL. When the proportion of corrected examples is 0\%---or even 25\% for some models---both methods yield equivalent performance, as the ``corrective'' task essentially reduces to replicating the previously predicted label. However, as the proportion of corrected examples increases, CICL's performance deteriorates further, highlighting the model's struggle to integrate feedback from corrected examples effectively.

Our findings suggest that the corrective nature of CICL introduces confusion rather than guiding the model towards improved predictions. Swapping (correcting) labels in the few-shot examples appears to disrupt the model’s internal representations, making it harder to generalize and refine predictions, especially when encountering harder examples.

\subsection{Statistical Analyses}

To assess the significance of these performance differences, we conduct statistical tests. A Shapiro-Wilk test for normality reveals no statistical evidence supporting normality in the distributions of ICL and CICL results. Therefore, we employ non-parametric tests to evaluate the impact of correction proportions on performance.

\begin{table}
    \centering
    \begin{tabular}{lrr}
        \toprule
        Corrected Proportion & Statistic & P-value \\
        \midrule
        0\% & 178.00 & 0.7916 \\
        25\% & 9453.00 & 4e-03* \\
        50\% & 20439.00 & 7e-04* \\
        75\% & 13553.00 & 1e-16* \\
        100\% & 984.00 & 9e-54* \\
        \bottomrule
    \end{tabular}
    \caption{Wilcoxon signed-rank test results for different correction proportions. * indicates significant differences at the 0.01 level.}
    \label{tab:stats_wilcoxon}
\end{table}

A Wilcoxon signed-rank test is used to determine the threshold at which ICL becomes statistically superior to CICL. As shown in Table \ref{tab:stats_wilcoxon}, from a correction proportion of 25\% onward, ICL demonstrates statistically significant superiority over CICL, reinforcing our conclusion that CICL's self-corrective mechanism is ineffective.

Additionally, a Kruskal-Wallis test is conducted to examine variation in performance across correction proportions (for CICL) and example conditions (for ICL). Table \ref{tab:stats_kruskal} highlights significant variation in CICL performance as the proportion of corrected examples increases, whereas ICL shows minimal variation across differing example conditions. These results underscore the robustness of ICL across diverse contexts, contrasting with CICL's instability under varying correction levels.

\begin{table}
    \centering
    \begin{tabular}{lrr}
        \toprule
        Method & Statistic & P-value \\
        \midrule
        ICL & 12.18 & 0.02 \\
        CICL & 677.54 & 3e-145* \\
        \bottomrule
    \end{tabular}
    \caption{Kruskal-Wallis test results for ICL and CICL. * indicates significant differences at the 0.01 level.}
    \label{tab:stats_kruskal}
\end{table}

\section{Impact of Harder Examples on ICL}

Although not the primary objective of this study, an intriguing finding emerges from our results. As shown in Figure \ref{fig:results}, increasing the proportion of incorrectly classified examples (i.e., higher ``corrected proportions'') in the few-shot context does not improve the performance of standard ICL. In fact, for some models, performance slightly declines as the corrected proportion increases.

Intuitively, one might expect that presenting the model with harder examples---those it initially misclassified---would enhance performance. The rationale is that such examples could help refine the model's decision boundaries and establish clearer classification thresholds, improving overall accuracy. However, our results do not support this.

When examining the performance trends of standard ICL (blue lines in Figure \ref{fig:results}), we observe that including harder examples in the few-shot context fails to yield any consistent improvement. In some cases, performance remains unchanged or even decreases slightly. Furthermore, statistical analysis using the Kruskal-Wallis test reveals no significant variation in standard ICL performance across different corrected proportions (Table \ref{tab:stats_kruskal}, first line). These findings suggest that simply exposing the model to harder examples does not provide the anticipated benefits for few-shot classification tasks.

This observation warrants further investigation into how example difficulty interacts with ICL, particularly in understanding why harder examples fail to contribute to improved model calibration or decision-making in this context.

\section{Conclusion}

In this paper, we introduced CICL, a novel approach aimed at leveraging incorrect model predictions to improve performance through self-correction. By including misclassified examples along with their correct labels in the prompt, CICL sought to refine the model's predictions in text classification tasks. However, contrary to our initial hypothesis, CICL consistently underperformed ICL across all models and datasets. Our findings revealed that the corrective nature of CICL often led to confusion rather than improvement. The swapping of labels in the few-shot examples disrupted the model's understanding of the task, resulting in degraded performance as the proportion of corrected examples increased.

Additionally, we explored an auxiliary finding regarding the impact of harder few-shot examples on standard ICL performance. Despite the expectation that presenting harder, misclassified examples could enhance the model's decision-making, our results showed no significant improvement across varying proportions of examples that required correction. This challenges the assumption that harder examples inherently contribute to better generalization in few-shot learning.

Overall, our study highlights the challenges of incorporating self-corrective mechanisms into LLMs through ICL and demonstrates that harder examples are not necessarily more useful than easier ones in standard ICL. These findings provide valuable insights for the development of more robust and effective few-shot learning methods with LLMs.

\section*{Limitations}
Despite the intuitive appeal of CICL, our findings indicate that it performs worse than standard ICL for text classification tasks. Several factors may contribute to this result.
First, our experiments are restricted to small-scale open-source LLMs due to computational constraints. Larger models may exhibit stronger reasoning and adaptation capabilities, potentially improving CICL performance. While preliminary experiments with the 70B version of Llama-3.1 yielded similar results to the smaller models, the impact of even larger models remains an open question.
Second, CICL may be more effective for tasks requiring multi-step reasoning, e.g., via chain-of-thought prompting, where the model can benefit from explicit corrective feedback to refine intermediate steps.
Third, LLMs are highly sensitive to prompt design. It is possible that alternative prompt formats or different ways of structuring corrective feedback could lead to better results.

\section*{Acknowledgments}
The research in this paper was funded by the Carl Zeiss Foundation, grant number P2022-08-009 (MAINCE project).

\bibliography{refs}

\begin{thebibliography}{26}
\providecommand{\natexlab}[1]{#1}

\bibitem[{Barbieri et~al.(2020)Barbieri, Camacho-Collados, Espinosa~Anke, and Neves}]{tweeteval_dataset}
Francesco Barbieri, Jose Camacho-Collados, Luis Espinosa~Anke, and Leonardo Neves. 2020.
\newblock \href {https://doi.org/10.18653/v1/2020.findings-emnlp.148} {{T}weet{E}val: Unified benchmark and comparative evaluation for tweet classification}.
\newblock In \emph{Findings of the Association for Computational Linguistics: EMNLP 2020}, pages 1644--1650, Online. Association for Computational Linguistics.

\bibitem[{Brown et~al.(2020)Brown, Mann, Ryder, Subbiah, Kaplan, Dhariwal, Neelakantan, Shyam, Sastry, Askell et~al.}]{brown2020fewshot}
Tom Brown, Benjamin Mann, Nick Ryder, Melanie Subbiah, Jared~D Kaplan, Prafulla Dhariwal, Arvind Neelakantan, Pranav Shyam, Girish Sastry, Amanda Askell, et~al. 2020.
\newblock \href {https://proceedings.neurips.cc/paper_files/paper/2020/file/1457c0d6bfcb4967418bfb8ac142f64a-Paper.pdf} {Language models are few-shot learners}.
\newblock In \emph{Advances in Neural Information Processing Systems}, volume~33, pages 1877--1901. Curran Associates, Inc.

\bibitem[{Dubey et~al.(2024)Dubey, Jauhri, Pandey, Kadian, Al-Dahle, Letman, Mathur, Schelten, Yang, Fan et~al.}]{dubey2024llama3}
Abhimanyu Dubey, Abhinav Jauhri, Abhinav Pandey, Abhishek Kadian, Ahmad Al-Dahle, Aiesha Letman, Akhil Mathur, Alan Schelten, Amy Yang, Angela Fan, et~al. 2024.
\newblock \href {https://arxiv.org/abs/2407.21783} {{The Llama 3 Herd of Models}}.
\newblock \emph{Preprint}, arXiv:2407.21783.

\bibitem[{Fei et~al.(2023)Fei, Hou, Chen, and Bosselut}]{fei2023domaincalibration}
Yu~Fei, Yifan Hou, Zeming Chen, and Antoine Bosselut. 2023.
\newblock \href {https://doi.org/10.18653/v1/2023.acl-long.783} {Mitigating label biases for in-context learning}.
\newblock In \emph{Proceedings of the 61st Annual Meeting of the Association for Computational Linguistics (Volume 1: Long Papers)}, pages 14014--14031, Toronto, Canada. Association for Computational Linguistics.

\bibitem[{Hu and Liu(2004)}]{cr_dataset}
Minqing Hu and Bing Liu. 2004.
\newblock \href {https://doi.org/10.1145/1014052.1014073} {Mining and summarizing customer reviews}.
\newblock In \emph{Proceedings of the Tenth ACM SIGKDD International Conference on Knowledge Discovery and Data Mining}, KDD '04, page 168–177, New York, NY, USA. Association for Computing Machinery.

\bibitem[{Huang et~al.(2023)Huang, Gu, Hou, Wu, Wang, Yu, and Han}]{huang2023selfimprove}
Jiaxin Huang, Shixiang Gu, Le~Hou, Yuexin Wu, Xuezhi Wang, Hongkun Yu, and Jiawei Han. 2023.
\newblock \href {https://doi.org/10.18653/v1/2023.emnlp-main.67} {Large language models can self-improve}.
\newblock In \emph{Proceedings of the 2023 Conference on Empirical Methods in Natural Language Processing}, pages 1051--1068, Singapore. Association for Computational Linguistics.

\bibitem[{Jiang et~al.(2023)Jiang, Sablayrolles, Mensch, Bamford, Chaplot, de~las Casas, Bressand, Lengyel, Lample, Saulnier et~al.}]{jiang2023mistral7b}
Albert~Q. Jiang, Alexandre Sablayrolles, Arthur Mensch, Chris Bamford, Devendra~Singh Chaplot, Diego de~las Casas, Florian Bressand, Gianna Lengyel, Guillaume Lample, Lucile Saulnier, et~al. 2023.
\newblock \href {https://arxiv.org/abs/2310.06825} {Mistral 7b}.
\newblock \emph{Preprint}, arXiv:2310.06825.

\bibitem[{Kumar et~al.(2024)Kumar, Zhuang, Agarwal, Su, Co-Reyes, Singh, Baumli, Iqbal, Bishop, Roelofs, Zhang, McKinney, Shrivastava, Paduraru, Tucker, Precup, Behbahani, and Faust}]{kumar2024rl}
Aviral Kumar, Vincent Zhuang, Rishabh Agarwal, Yi~Su, John~D Co-Reyes, Avi Singh, Kate Baumli, Shariq Iqbal, Colton Bishop, Rebecca Roelofs, Lei~M Zhang, Kay McKinney, Disha Shrivastava, Cosmin Paduraru, George Tucker, Doina Precup, Feryal Behbahani, and Aleksandra Faust. 2024.
\newblock \href {https://arxiv.org/abs/2409.12917} {Training language models to self-correct via reinforcement learning}.
\newblock \emph{Preprint}, arXiv:2409.12917.

\bibitem[{Lester et~al.(2021)Lester, Al-Rfou, and Constant}]{lester2021power}
Brian Lester, Rami Al-Rfou, and Noah Constant. 2021.
\newblock \href {https://doi.org/10.18653/v1/2021.emnlp-main.243} {The power of scale for parameter-efficient prompt tuning}.
\newblock In \emph{Proceedings of the 2021 Conference on Empirical Methods in Natural Language Processing}, pages 3045--3059, Online and Punta Cana, Dominican Republic. Association for Computational Linguistics.

\bibitem[{Lhoest et~al.(2021)Lhoest, Villanova~del Moral, Jernite, Thakur, von Platen, Patil, Chaumond, Drame, Plu, Tunstall et~al.}]{huggingface2021datasets}
Quentin Lhoest, Albert Villanova~del Moral, Yacine Jernite, Abhishek Thakur, Patrick von Platen, Suraj Patil, Julien Chaumond, Mariama Drame, Julien Plu, Lewis Tunstall, et~al. 2021.
\newblock \href {https://doi.org/10.18653/v1/2021.emnlp-demo.21} {Datasets: A community library for natural language processing}.
\newblock In \emph{Proceedings of the 2021 Conference on Empirical Methods in Natural Language Processing: System Demonstrations}, pages 175--184, Online and Punta Cana, Dominican Republic. Association for Computational Linguistics.

\bibitem[{Li and Liang(2021)}]{li2021prefix}
Xiang~Lisa Li and Percy Liang. 2021.
\newblock \href {https://doi.org/10.18653/v1/2021.acl-long.353} {Prefix-tuning: Optimizing continuous prompts for generation}.
\newblock In \emph{Proceedings of the 59th Annual Meeting of the Association for Computational Linguistics and the 11th International Joint Conference on Natural Language Processing (Volume 1: Long Papers)}, pages 4582--4597, Online. Association for Computational Linguistics.

\bibitem[{Liu et~al.(2022)Liu, Shen, Zhang, Dolan, Carin, and Chen}]{liu2022gpt3examples}
Jiachang Liu, Dinghan Shen, Yizhe Zhang, Bill Dolan, Lawrence Carin, and Weizhu Chen. 2022.
\newblock \href {https://doi.org/10.18653/v1/2022.deelio-1.10} {What makes good in-context examples for {GPT}-3?}
\newblock In \emph{Proceedings of Deep Learning Inside Out (DeeLIO 2022): The 3rd Workshop on Knowledge Extraction and Integration for Deep Learning Architectures}, pages 100--114, Dublin, Ireland and Online. Association for Computational Linguistics.

\bibitem[{Lu et~al.(2022)Lu, Bartolo, Moore, Riedel, and Stenetorp}]{lu2022fantastically}
Yao Lu, Max Bartolo, Alastair Moore, Sebastian Riedel, and Pontus Stenetorp. 2022.
\newblock \href {https://doi.org/10.18653/v1/2022.acl-long.556} {Fantastically ordered prompts and where to find them: Overcoming few-shot prompt order sensitivity}.
\newblock In \emph{Proceedings of the 60th Annual Meeting of the Association for Computational Linguistics (Volume 1: Long Papers)}, pages 8086--8098, Dublin, Ireland. Association for Computational Linguistics.

\bibitem[{Madaan et~al.(2023)Madaan, Tandon, Gupta, Hallinan, Gao, Wiegreffe, Alon, Dziri, Prabhumoye, Yang, Gupta, Majumder, Hermann, Welleck, Yazdanbakhsh, and Clark}]{madaan2023selfrefine}
Aman Madaan, Niket Tandon, Prakhar Gupta, Skyler Hallinan, Luyu Gao, Sarah Wiegreffe, Uri Alon, Nouha Dziri, Shrimai Prabhumoye, Yiming Yang, Shashank Gupta, Bodhisattwa~Prasad Majumder, Katherine Hermann, Sean Welleck, Amir Yazdanbakhsh, and Peter Clark. 2023.
\newblock \href {https://openreview.net/forum?id=S37hOerQLB} {Self-refine: Iterative refinement with self-feedback}.
\newblock In \emph{Thirty-seventh Conference on Neural Information Processing Systems}.

\bibitem[{Malo et~al.(2014)Malo, Sinha, Korhonen, Wallenius, and Takala}]{financial_phrasebank_dataset}
Pekka Malo, Ankur Sinha, Pekka Korhonen, Jyrki Wallenius, and Pyry Takala. 2014.
\newblock Good debt or bad debt: Detecting semantic orientations in economic texts.
\newblock \emph{Journal of the Association for Information Science and Technology}, 65(4):782--796.

\bibitem[{Min et~al.(2022)Min, Lyu, Holtzman, Artetxe, Lewis, Hajishirzi, and Zettlemoyer}]{min2022rethinking}
Sewon Min, Xinxi Lyu, Ari Holtzman, Mikel Artetxe, Mike Lewis, Hannaneh Hajishirzi, and Luke Zettlemoyer. 2022.
\newblock \href {https://doi.org/10.18653/v1/2022.emnlp-main.759} {Rethinking the role of demonstrations: What makes in-context learning work?}
\newblock In \emph{Proceedings of the 2022 Conference on Empirical Methods in Natural Language Processing}, pages 11048--11064, Abu Dhabi, United Arab Emirates. Association for Computational Linguistics.

\bibitem[{Monea et~al.(2024)Monea, Bosselut, Brantley, and Artzi}]{monea2024icrl}
Giovanni Monea, Antoine Bosselut, Kianté Brantley, and Yoav Artzi. 2024.
\newblock \href {https://arxiv.org/abs/2410.05362v1} {Llms are in-context reinforcement learners}.
\newblock \emph{Preprint}, arXiv:2410.05362v1.

\bibitem[{Pang and Lee(2004)}]{subj_dataset}
Bo~Pang and Lillian Lee. 2004.
\newblock \href {https://doi.org/10.3115/1218955.1218990} {A sentimental education: Sentiment analysis using subjectivity summarization based on minimum cuts}.
\newblock In \emph{Proceedings of the 42nd Annual Meeting of the Association for Computational Linguistics ({ACL}-04)}, pages 271--278, Barcelona, Spain.

\bibitem[{Pang and Lee(2005)}]{mr_dataset}
Bo~Pang and Lillian Lee. 2005.
\newblock \href {https://doi.org/10.3115/1219840.1219855} {Seeing stars: Exploiting class relationships for sentiment categorization with respect to rating scales}.
\newblock In \emph{Proceedings of the 43rd Annual Meeting of the Association for Computational Linguistics ({ACL}{'}05)}, pages 115--124, Ann Arbor, Michigan. Association for Computational Linguistics.

\bibitem[{Qwen et~al.(2024)Qwen, Yang, Yang, Zhang, Hui, Zheng, Yu, Li, Liu, Huang et~al.}]{qwen2024qwen25}
Qwen, An~Yang, Baosong Yang, Beichen Zhang, Binyuan Hui, Bo~Zheng, Bowen Yu, Chengyuan Li, Dayiheng Liu, Fei Huang, et~al. 2024.
\newblock \href {https://arxiv.org/abs/2412.15115} {Qwen2.5 technical report}.
\newblock \emph{Preprint}, arXiv:2412.15115.

\bibitem[{Sheng and Uthus(2020)}]{poem_sentiment_dataset}
Emily Sheng and David Uthus. 2020.
\newblock \href {https://aclanthology.org/2020.gebnlp-1.9} {Investigating societal biases in a poetry composition system}.
\newblock In \emph{Proceedings of the Second Workshop on Gender Bias in Natural Language Processing}, pages 93--106, Barcelona, Spain (Online). Association for Computational Linguistics.

\bibitem[{Socher et~al.(2013)Socher, Perelygin, Wu, Chuang, Manning, Ng, and Potts}]{sst_dataset}
Richard Socher, Alex Perelygin, Jean Wu, Jason Chuang, Christopher~D. Manning, Andrew Ng, and Christopher Potts. 2013.
\newblock \href {https://www.aclweb.org/anthology/D13-1170} {Recursive deep models for semantic compositionality over a sentiment treebank}.
\newblock In \emph{Proceedings of the 2013 Conference on Empirical Methods in Natural Language Processing}, pages 1631--1642, Seattle, Washington, USA. Association for Computational Linguistics.

\bibitem[{Voorhees and Tice(2000)}]{trec_dataset}
Ellen~M. Voorhees and Dawn~M. Tice. 2000.
\newblock \href {https://doi.org/10.1145/345508.345577} {Building a question answering test collection}.
\newblock In \emph{Proceedings of the 23rd Annual International ACM SIGIR Conference on Research and Development in Information Retrieval}, SIGIR '00, page 200–207, New York, NY, USA. Association for Computing Machinery.

\bibitem[{Wang and Komatsuzaki(2021)}]{wang2021gptj}
Ben Wang and Aran Komatsuzaki. 2021.
\newblock {GPT-J-6B: A 6 Billion Parameter Autoregressive Language Model}.
\newblock \url{https://github.com/kingoflolz/mesh-transformer-jax}.

\bibitem[{Zhang et~al.(2015)Zhang, Zhao, and LeCun}]{agnews_dbpedia_dataset}
Xiang Zhang, Junbo Zhao, and Yann LeCun. 2015.
\newblock \href {https://proceedings.neurips.cc/paper_files/paper/2015/file/250cf8b51c773f3f8dc8b4be867a9a02-Paper.pdf} {Character-level convolutional networks for text classification}.
\newblock In \emph{Advances in Neural Information Processing Systems}, volume~28. Curran Associates, Inc.

\bibitem[{Zhao et~al.(2021)Zhao, Wallace, Feng, Klein, and Singh}]{zhao2021calibrate}
Zihao Zhao, Eric Wallace, Shi Feng, Dan Klein, and Sameer Singh. 2021.
\newblock \href {https://proceedings.mlr.press/v139/zhao21c.html} {Calibrate before use: Improving few-shot performance of language models}.
\newblock In \emph{Proceedings of the 38th International Conference on Machine Learning}, volume 139 of \emph{Proceedings of Machine Learning Research}, pages 12697--12706. PMLR.

\end{thebibliography}
\label{sec:bibtex}

\appendix

\section{Datasets Details}
\label{app:datasets}

Our experiments utilize 17 datasets spanning various text classification tasks, all of which are commonly used in prior research \cite{min2022rethinking, lu2022fantastically, zhao2021calibrate, fei2023domaincalibration}. All datasets are accessed through the HuggingFace Datasets library \cite{huggingface2021datasets}. For evaluation, we use the provided test sets when available. If no test set is provided, we create a stratified development set by sampling from the training data, ensuring the class distribution is preserved. Detailed dataset information, including task type, class counts, and data distribution, is summarized in Table \ref{tab:full_datasets}.

As shown in Table \ref{tab:full_datasets}, most datasets are imbalanced. To account for this, we use macro F1 as the evaluation metric, as it equally weighs all classes and ensures a fair assessment of the model's performance across both frequent and rare classes.

\begin{table*}[]
    \centering
    \begin{tabular}{lcc}
        \toprule
        Dataset & \# Classes & Balanced \\
        \midrule
        \multicolumn{3}{l}{\textit{Sentiment and topic classification}} \\
        SST-2 \cite{sst_dataset} & 2 & \ding{51} \\
        SST-5 \cite{sst_dataset} & 5 & \ding{55} \\
        MR \cite{mr_dataset} & 2 & \ding{51} \\
        CR \cite{cr_dataset} & 2 & \ding{51} \\
        financial\_phrasebank \cite{financial_phrasebank_dataset} & 3 & \ding{55} \\
        poem\_sentiment \cite{poem_sentiment_dataset} & 4 & \ding{55} \\
        Subj \cite{subj_dataset} & 2 & \ding{55} \\
        AG News \cite{agnews_dbpedia_dataset} & 4 & \ding{51} \\
        DBpedia \cite{agnews_dbpedia_dataset} & 14 & \ding{51} \\
        TREC \cite{trec_dataset} & 6 & \ding{55} \\
        \midrule
        \multicolumn{3}{l}{\textit{Detection}} \\
        tweet\_eval-hate \cite{tweeteval_dataset} & 2 & \ding{55} \\
        tweet\_eval-irony \cite{tweeteval_dataset} & 2 & \ding{55} \\
        tweet\_eval-offensive \cite{tweeteval_dataset} & 2 & \ding{55} \\
        tweet\_eval-stance\_atheism \cite{tweeteval_dataset} & 3 & \ding{55} \\
        tweet\_eval-stance\_feminist \cite{tweeteval_dataset} & 3 & \ding{55} \\
        hate\_speech18 \cite{tweeteval_dataset} & 2 & \ding{55} \\
        ethos-binary \cite{tweeteval_dataset} & 2 & \ding{55} \\
        \bottomrule
    \end{tabular}
    \caption{Full dataset information.}
    \label{tab:full_datasets}
\end{table*}

\section{Prompt Formats}
\label{app:prompts}

\subsection{In-Context Learning Prompt}
\label{app:icl_prompt}

\begin{figure}[H]
    \centering
    \fbox{%
    \ttfamily
    \begin{tabular}{l}
        Text: \{example\_1\} \\
        Label: \{ground\_truth\_1\} \\[1ex]
        Text: \{example\_2\} \\
        Label: \{ground\_truth\_2\} \\
        ... \\
        Text: \{example\_$k$\} \\
        Label: \{ground\_truth\_$k$\} \\[1ex]
        Text: \{input\_text\} \\
        Label:
    \end{tabular}%
    }
    \caption{Prompt format for standard ICL, showing ground truth labels for $k$ examples.}
    \label{fig:icl_prompt}
\end{figure}

\subsubsection*{Example}

Below is an example of a standard ICL prompt for the TREC dataset with $k=8$ few-shot examples.\\

\noindent
\texttt{
    Text: What is the name of the tallest mountain in the world? \\
    Label: location \\[1ex]
    Text: How many eyes does a bat have? \\
    Label: numeric \\[1ex]
    Text: What does Ms., Miss, and Mrs. stand for? \\
    Label: abbreviation \\[1ex]
    Text: What does IQ stand for? \\
    Label: abbreviation \\[1ex]
    Text: What were the achievements of Richard Nixon? \\
    Label: entity \\[1ex]
    Text: What is the C programming language? \\
    Label: description \\[1ex]
    Text: Who was considered to be the father of psychology? \\
    Label: human \\[1ex]
    Text: What are the top five oil-producing countries in the world? \\
    Label: location \\[1ex]
    Text: What are the stars made of? \\
    Label:
}

\subsection{Corrective In-Context Learning Prompt}
\label{app:cicl_prompt}

\begin{figure}[H]
    \centering
    \fbox{
    \ttfamily
    \begin{tabular}{l}
        Text: \{example\_1\} \\
        Predicted label: \{predicted\_label\_1\} \\
        Correct label: \{ground\_truth\_1\} \\[1ex]
        Text: \{example\_2\} \\
        Predicted label: \{predicted\_label\_2\} \\
        Correct label: \{ground\_truth\_2\} \\
        ... \\
        Text: \{example\_$k$\} \\
        Predicted label: \{predicted\_label\_$k$\} \\
        Correct label: \{ground\_truth\_$k$\} \\[1ex]
        Text: \{input\_text\} \\
        Predicted label: \{predicted\_label\} \\
        Correct label:
    \end{tabular}%
    }
    \caption{Prompt format for CICL, showing predicted and ground truth (``correct'') labels for $k$ examples.}
    \label{fig:cicl_prompt}
\end{figure}

\subsubsection*{Example}

Below is an example of a CICL prompt for the TREC dataset. There are $k=8$ few-shot examples, with 50\% being corrected examples. This means 4 examples are correctly predicted by ICL (in positions 2, 4, 6, and 7), and 4 are corrected examples (in positions 1, 3, 5, and 8). The final instance represents the current input, where ICL predicted the label ``description.'' The task of the LLM is to predict the correct label based on the input text, the predicted label, and the corrections made in previous examples. The true label for this instance is ``entity,'' and the goal of CICL is to make this correction.\\

\noindent
\texttt{
    Text: What is the name of the tallest mountain in the world? \\
    Predicted label: entity \\
    Correct label: location \\[1ex]
    Text: How many eyes does a bat have? \\
    Predicted label: numeric \\
    Correct label: numeric \\[1ex]
    Text: What does Ms., Miss, and Mrs. stand for? \\
    Predicted label: description \\
    Correct label: abbreviation \\[1ex]
    Text: What does IQ stand for? \\
    Predicted label: abbreviation \\
    Correct label: abbreviation \\[1ex]
    Text: What were the achievements of Richard Nixon? \\
    Predicted label: human \\
    Correct label: entity \\[1ex]
    Text: What is the C programming language? \\
    Predicted label: description \\
    Correct label: description \\[1ex]
    Text: Who was considered to be the father of psychology? \\
    Predicted label: human \\
    Correct label: human \\[1ex]
    Text: What are the top five oil-producing countries in the world? \\
    Predicted label: numeric \\
    Correct label: location \\[1ex]
    Text: What are the stars made of? \\
    Predicted label: description \\
    Correct label:
}

\section{Detailed Results}
\label{app:detailed_results}

Table \ref{tab:detailed_results} presents the detailed results for all combinations of models, datasets, corrected example proportions, and approaches (ICL and CICL). Each configuration is evaluated five times using different random seeds to ensure diverse selections of examples, minimizing the impact of randomness on the results. The table reports the mean and standard deviation for each configuration, providing a comprehensive view of the performance across variations.

\begin{table*}[ht!]
\tiny
\centering
\setlength{\tabcolsep}{4pt}
\begin{tabular}{|l|c||c|c||c|c||c|c||c|c|c|}
\hline
\multirow{2}{*}{\textbf{Dataset}} & \multirow{2}{*}{\shortstack{\textbf{Corrected}\\\textbf{Proportion}}} & \multicolumn{2}{c||}{\textbf{Llama-3.1}} & \multicolumn{2}{c||}{\textbf{Mistral 7B}} & \multicolumn{2}{c||}{\textbf{Qwen2.5}} & \multicolumn{2}{c|}{\textbf{GPT-J}} \\
\cline{3-10}
 & & \textbf{ICL} & \textbf{CICL} & \textbf{ICL} & \textbf{CICL} & \textbf{ICL} & \textbf{CICL} & \textbf{ICL} & \textbf{CICL} \\
\hline\hline
\multirow{5}{*}{AG News} & 0\% & $87.3_{1.6}$ & $87.3_{1.6}$ & $85.5_{1.0}$ & $85.5_{1.0}$ & $83.4_{1.7}$ & $83.6_{1.6}$ & $73.9_{4.4}$ & $73.9_{4.4}$ \\
 & 25\% & $83.7_{4.0}$ & $84.0_{3.6}$ & $86.6_{1.0}$ & $86.5_{0.7}$ & $85.3_{1.1}$ & $85.7_{1.2}$ & $74.2_{0.8}$ & $74.2_{0.8}$ \\
 & 50\% & $84.1_{3.4}$ & $83.0_{4.4}$ & $85.0_{2.3}$ & $76.4_{13.0}$ & $83.2_{2.4}$ & $82.0_{2.4}$ & $73.9_{7.3}$ & $70.5_{7.8}$ \\
 & 75\% & $68.7_{20.2}$ & $56.0_{24.0}$ & $80.9_{3.9}$ & $47.6_{22.8}$ & $80.0_{6.1}$ & $70.4_{8.7}$ & $75.9_{2.4}$ & $47.6_{7.3}$ \\
 & 100\% & $83.6_{2.8}$ & $13.5_{2.1}$ & $85.0_{3.2}$ & $12.2_{2.2}$ & $85.1_{1.9}$ & $15.3_{5.1}$ & $77.1_{3.8}$ & $14.6_{2.0}$ \\
\hline
\multirow{5}{*}{CR} & 0\% & $93.2_{0.5}$ & $93.2_{0.5}$ & $93.2_{0.4}$ & $93.2_{0.4}$ & $93.2_{0.5}$ & $93.1_{0.5}$ & $87.6_{2.0}$ & $87.6_{2.0}$ \\
 & 25\% & $92.0_{1.3}$ & $91.8_{1.5}$ & $92.0_{0.7}$ & $91.5_{0.7}$ & $91.8_{1.8}$ & $92.0_{0.8}$ & $87.8_{2.7}$ & $84.8_{7.9}$ \\
 & 50\% & $93.5_{0.7}$ & $83.6_{7.3}$ & $91.7_{1.9}$ & $89.5_{1.0}$ & $91.4_{1.4}$ & $91.5_{0.9}$ & $87.1_{2.3}$ & $84.5_{5.4}$ \\
 & 75\% & $91.1_{1.9}$ & $55.7_{18.9}$ & $87.2_{6.7}$ & $71.8_{14.1}$ & $89.2_{5.7}$ & $75.3_{18.2}$ & $84.3_{5.6}$ & $58.2_{20.3}$ \\
 & 100\% & $91.1_{1.0}$ & $8.0_{0.8}$ & $91.4_{1.0}$ & $7.8_{0.6}$ & $84.3_{7.5}$ & $14.9_{5.3}$ & $86.2_{5.8}$ & $11.9_{4.4}$ \\
\hline
\multirow{5}{*}{DBpedia} & 0\% & $85.6_{1.4}$ & $85.6_{1.4}$ & $82.6_{2.4}$ & $82.6_{2.4}$ & $75.7_{2.2}$ & $75.8_{2.1}$ & $77.4_{1.0}$ & $77.4_{1.0}$ \\
 & 25\% & $84.0_{3.0}$ & $84.2_{2.9}$ & $80.7_{0.9}$ & $81.0_{1.3}$ & $75.5_{1.4}$ & $76.9_{1.3}$ & $75.1_{1.6}$ & $74.7_{2.2}$ \\
 & 50\% & $80.5_{1.1}$ & $84.6_{1.8}$ & $76.2_{7.1}$ & $79.1_{5.3}$ & $74.1_{1.4}$ & $81.4_{6.0}$ & $75.1_{2.4}$ & $73.3_{4.7}$ \\
 & 75\% & $81.6_{1.1}$ & $86.6_{3.2}$ & $78.0_{1.0}$ & $79.3_{4.6}$ & $71.2_{1.9}$ & $74.7_{6.3}$ & $73.8_{1.7}$ & $59.0_{11.2}$ \\
 & 100\% & $78.8_{2.1}$ & $12.8_{3.4}$ & $66.1_{11.8}$ & $9.6_{4.4}$ & $68.8_{3.4}$ & $13.6_{4.8}$ & $72.8_{2.6}$ & $12.3_{3.8}$ \\
\hline
\multirow{5}{*}{ethos-binary} & 0\% & $72.8_{3.3}$ & $72.8_{3.3}$ & $79.6_{3.6}$ & $79.6_{3.6}$ & $80.6_{2.2}$ & $80.6_{2.2}$ & $56.9_{9.5}$ & $56.6_{9.3}$ \\
 & 25\% & $63.2_{10.8}$ & $62.7_{10.7}$ & $76.1_{5.7}$ & $73.1_{4.6}$ & $77.7_{4.1}$ & $77.9_{3.9}$ & $44.6_{15.0}$ & $45.3_{14.6}$ \\
 & 50\% & $67.6_{4.6}$ & $62.6_{7.8}$ & $69.2_{7.0}$ & $60.8_{17.6}$ & $77.5_{3.0}$ & $77.8_{1.8}$ & $58.7_{8.4}$ & $57.4_{10.9}$ \\
 & 75\% & $55.6_{16.2}$ & $46.7_{15.7}$ & $70.6_{12.9}$ & $55.0_{14.8}$ & $78.9_{3.7}$ & $71.8_{9.8}$ & $55.8_{14.4}$ & $53.1_{18.8}$ \\
 & 100\% & $51.4_{7.3}$ & $30.8_{2.8}$ & $64.3_{13.8}$ & $26.7_{4.5}$ & $77.9_{2.3}$ & $23.8_{3.4}$ & $36.3_{0.0}$ & $30.1_{0.0}$ \\
\hline
\multirow{5}{*}{financial\_phrasebank} & 0\% & $77.9_{8.3}$ & $77.9_{8.3}$ & $84.1_{0.9}$ & $84.1_{0.9}$ & $85.4_{3.1}$ & $85.2_{3.0}$ & $59.5_{12.0}$ & $59.5_{12.0}$ \\
 & 25\% & $78.6_{4.8}$ & $78.7_{4.8}$ & $82.5_{1.2}$ & $82.4_{1.3}$ & $82.4_{4.3}$ & $82.4_{2.9}$ & $58.0_{12.0}$ & $57.6_{12.3}$ \\
 & 50\% & $74.9_{9.3}$ & $68.8_{8.5}$ & $83.5_{1.3}$ & $83.3_{1.7}$ & $84.2_{1.2}$ & $83.6_{1.8}$ & $61.9_{11.1}$ & $61.8_{9.7}$ \\
 & 75\% & $79.4_{8.8}$ & $43.4_{10.4}$ & $82.2_{3.0}$ & $70.3_{11.9}$ & $84.7_{1.5}$ & $82.3_{2.5}$ & $50.2_{8.3}$ & $17.3_{6.0}$ \\
 & 100\% & $81.1_{5.5}$ & $10.5_{4.3}$ & $78.5_{5.3}$ & $25.9_{7.8}$ & $84.1_{4.0}$ & $23.2_{14.9}$ & $49.6_{13.8}$ & $13.8_{7.9}$ \\
\hline
\multirow{5}{*}{hate\_speech18} & 0\% & $53.5_{4.1}$ & $53.5_{4.1}$ & $67.5_{1.8}$ & $67.5_{1.8}$ & $68.4_{2.1}$ & $68.3_{2.1}$ & $38.9_{13.1}$ & $38.7_{12.9}$ \\
 & 25\% & $47.9_{8.8}$ & $47.4_{8.6}$ & $57.2_{6.2}$ & $51.3_{13.0}$ & $66.0_{2.5}$ & $65.7_{2.6}$ & $37.2_{10.0}$ & $34.4_{10.8}$ \\
 & 50\% & $46.4_{11.1}$ & $46.1_{10.7}$ & $62.2_{5.0}$ & $60.9_{5.7}$ & $63.4_{3.3}$ & $64.9_{2.1}$ & $38.1_{4.6}$ & $37.5_{4.6}$ \\
 & 75\% & $36.9_{5.5}$ & $39.8_{12.5}$ & $52.8_{14.0}$ & $49.2_{18.1}$ & $64.2_{3.5}$ & $58.3_{7.3}$ & $37.5_{11.7}$ & $42.9_{17.2}$ \\
 & 100\% & $42.8_{13.9}$ & $34.3_{6.8}$ & $57.8_{4.7}$ & $24.2_{4.9}$ & $63.6_{2.1}$ & $21.4_{4.6}$ & $45.5_{3.0}$ & $15.8_{11.4}$ \\
\hline
\multirow{5}{*}{MR} & 0\% & $93.6_{0.5}$ & $93.6_{0.5}$ & $93.8_{0.4}$ & $93.8_{0.4}$ & $92.6_{0.7}$ & $92.6_{0.7}$ & $90.7_{1.0}$ & $90.7_{1.0}$ \\
 & 25\% & $93.1_{1.3}$ & $93.0_{1.0}$ & $94.1_{0.3}$ & $94.0_{0.4}$ & $93.0_{0.6}$ & $92.7_{0.6}$ & $88.6_{2.8}$ & $88.3_{2.6}$ \\
 & 50\% & $93.1_{0.9}$ & $89.8_{3.7}$ & $93.9_{0.6}$ & $92.7_{1.4}$ & $92.4_{0.5}$ & $91.0_{0.8}$ & $81.6_{8.4}$ & $81.8_{8.1}$ \\
 & 75\% & $91.7_{4.1}$ & $44.5_{8.3}$ & $91.6_{2.9}$ & $79.4_{10.4}$ & $92.6_{0.5}$ & $69.0_{14.0}$ & $76.0_{21.6}$ & $62.7_{15.7}$ \\
 & 100\% & $81.8_{11.3}$ & $14.8_{6.9}$ & $80.7_{21.7}$ & $13.3_{9.7}$ & $90.2_{4.2}$ & $9.8_{3.0}$ & $75.1_{19.1}$ & $17.7_{7.9}$ \\
\hline
\multirow{5}{*}{poem\_sentiment} & 0\% & $39.4_{3.6}$ & $39.4_{3.6}$ & $54.0_{5.6}$ & $54.0_{5.6}$ & $54.0_{3.5}$ & $54.0_{3.5}$ & $13.0_{4.2}$ & $13.0_{4.2}$ \\
 & 25\% & $42.2_{7.1}$ & $42.2_{7.1}$ & $53.4_{11.3}$ & $53.9_{10.3}$ & $50.9_{6.6}$ & $51.4_{6.3}$ & $20.7_{11.0}$ & $22.1_{11.1}$ \\
 & 50\% & $41.7_{12.8}$ & $40.6_{12.6}$ & $57.8_{6.1}$ & $57.2_{4.9}$ & $53.6_{5.9}$ & $54.5_{5.6}$ & $22.5_{8.0}$ & $28.0_{2.2}$ \\
 & 75\% & $47.2_{12.0}$ & $25.1_{9.0}$ & $49.5_{11.1}$ & $50.3_{9.7}$ & $51.5_{7.9}$ & $47.0_{6.1}$ & $40.4_{4.9}$ & $27.7_{12.5}$ \\
 & 100\% & $54.7_{8.6}$ & $9.8_{3.6}$ & $57.7_{11.6}$ & $17.8_{7.8}$ & $52.0_{7.0}$ & $21.6_{4.2}$ & $25.4_{11.7}$ & $28.5_{5.0}$ \\
\hline
\multirow{5}{*}{SST-2} & 0\% & $91.6_{1.3}$ & $91.6_{1.3}$ & $92.4_{1.5}$ & $92.4_{1.5}$ & $93.8_{1.0}$ & $93.8_{1.0}$ & $83.0_{9.0}$ & $83.0_{9.0}$ \\
 & 25\% & $91.2_{1.1}$ & $90.7_{1.8}$ & $91.1_{0.9}$ & $91.1_{0.8}$ & $91.9_{2.8}$ & $92.7_{2.6}$ & $72.5_{17.6}$ & $73.9_{18.0}$ \\
 & 50\% & $90.2_{1.5}$ & $86.3_{4.0}$ & $89.8_{2.1}$ & $89.0_{2.6}$ & $92.4_{1.1}$ & $94.5_{0.5}$ & $80.8_{8.4}$ & $80.7_{5.4}$ \\
 & 75\% & $85.2_{9.3}$ & $40.5_{8.6}$ & $86.1_{7.1}$ & $68.8_{21.6}$ & $92.0_{3.0}$ & $86.0_{5.6}$ & $68.9_{20.1}$ & $41.2_{23.6}$ \\
 & 100\% & $86.0_{5.0}$ & $12.9_{3.8}$ & $87.9_{4.5}$ & $11.4_{3.3}$ & $90.6_{1.1}$ & $9.9_{0.4}$ & $71.7_{7.8}$ & $21.0_{4.2}$ \\
\hline
\multirow{5}{*}{SST-5} & 0\% & $45.4_{2.4}$ & $45.4_{2.4}$ & $42.6_{4.3}$ & $42.6_{4.3}$ & $42.0_{2.1}$ & $42.0_{2.1}$ & $31.5_{9.8}$ & $31.5_{9.8}$ \\
 & 25\% & $43.7_{2.1}$ & $43.6_{1.5}$ & $39.0_{4.6}$ & $38.9_{4.7}$ & $41.5_{2.2}$ & $41.6_{2.2}$ & $27.4_{3.4}$ & $23.6_{7.2}$ \\
 & 50\% & $39.0_{4.4}$ & $40.0_{4.7}$ & $40.5_{4.3}$ & $39.8_{3.9}$ & $42.3_{4.2}$ & $38.1_{3.4}$ & $34.5_{6.3}$ & $31.9_{8.6}$ \\
 & 75\% & $43.3_{2.4}$ & $39.4_{3.9}$ & $38.8_{4.6}$ & $37.3_{2.7}$ & $39.1_{6.0}$ & $42.0_{3.5}$ & $33.2_{3.1}$ & $28.3_{7.2}$ \\
 & 100\% & $41.4_{2.2}$ & $26.3_{1.2}$ & $44.3_{4.1}$ & $31.9_{4.2}$ & $41.1_{6.6}$ & $29.6_{1.9}$ & $28.2_{4.5}$ & $20.9_{5.2}$ \\
\hline
\multirow{5}{*}{Subj} & 0\% & $87.1_{7.9}$ & $87.1_{7.9}$ & $75.3_{19.5}$ & $75.3_{19.5}$ & $79.7_{6.3}$ & $79.4_{6.3}$ & $43.0_{12.0}$ & $43.0_{12.0}$ \\
 & 25\% & $86.9_{3.3}$ & $86.0_{3.2}$ & $68.3_{13.3}$ & $62.4_{14.2}$ & $80.8_{3.7}$ & $70.6_{3.9}$ & $55.7_{18.3}$ & $42.9_{14.1}$ \\
 & 50\% & $83.6_{4.5}$ & $57.2_{13.2}$ & $59.0_{19.0}$ & $51.5_{21.2}$ & $82.1_{2.4}$ & $68.4_{3.6}$ & $57.2_{15.4}$ & $49.8_{11.4}$ \\
 & 75\% & $82.0_{6.8}$ & $25.0_{7.2}$ & $37.4_{4.9}$ & $54.8_{18.1}$ & $78.4_{6.8}$ & $47.8_{12.7}$ & $70.4_{3.6}$ & $44.3_{14.7}$ \\
 & 100\% & $82.4_{9.3}$ & $15.2_{5.9}$ & $47.9_{20.3}$ & $28.2_{8.1}$ & $84.4_{2.6}$ & $14.6_{1.9}$ & $39.9_{9.7}$ & $32.8_{1.2}$ \\
\hline
\multirow{5}{*}{TREC-6} & 0\% & $75.0_{7.3}$ & $75.0_{7.3}$ & $56.6_{7.6}$ & $56.6_{7.6}$ & $74.7_{3.6}$ & $74.7_{3.6}$ & $51.2_{7.7}$ & $51.2_{7.7}$ \\
 & 25\% & $74.6_{7.4}$ & $74.5_{7.7}$ & $56.9_{12.2}$ & $57.2_{12.2}$ & $81.6_{3.9}$ & $81.1_{3.6}$ & $46.8_{3.6}$ & $47.0_{3.3}$ \\
 & 50\% & $75.8_{3.5}$ & $75.6_{2.9}$ & $60.1_{4.9}$ & $59.4_{3.0}$ & $74.8_{5.0}$ & $77.1_{3.3}$ & $49.9_{7.1}$ & $49.4_{2.2}$ \\
 & 75\% & $77.7_{3.0}$ & $69.6_{8.4}$ & $49.9_{16.4}$ & $52.2_{6.7}$ & $81.7_{1.9}$ & $63.9_{10.3}$ & $48.7_{7.7}$ & $44.7_{9.9}$ \\
 & 100\% & $66.8_{9.0}$ & $13.4_{6.0}$ & $50.6_{13.6}$ & $22.6_{7.2}$ & $77.0_{2.6}$ & $9.1_{4.3}$ & $42.9_{6.4}$ & $20.9_{1.5}$ \\
\hline
\multirow{5}{*}{tweet\_eval\_atheism} & 0\% & $21.2_{10.1}$ & $21.2_{10.1}$ & $27.6_{10.3}$ & $27.6_{10.3}$ & $34.0_{4.3}$ & $34.0_{4.3}$ & $12.1_{1.3}$ & $12.1_{1.3}$ \\
 & 25\% & $26.6_{4.8}$ & $28.3_{3.1}$ & $33.2_{11.5}$ & $33.2_{11.5}$ & $37.1_{5.9}$ & $37.7_{5.6}$ & $20.1_{10.6}$ & $21.5_{7.7}$ \\
 & 50\% & $31.5_{9.0}$ & $34.4_{8.1}$ & $33.7_{6.5}$ & $35.8_{8.0}$ & $40.9_{7.9}$ & $37.9_{6.0}$ & $16.6_{5.6}$ & $23.1_{10.7}$ \\
 & 75\% & $42.5_{7.0}$ & $38.4_{10.8}$ & $37.4_{7.3}$ & $30.0_{7.1}$ & $36.6_{11.7}$ & $36.6_{11.1}$ & $18.4_{6.1}$ & $23.0_{4.3}$ \\
 & 100\% & $41.0_{8.1}$ & $22.5_{4.9}$ & $39.0_{10.9}$ & $16.8_{3.6}$ & $35.2_{8.1}$ & $23.5_{3.2}$ & $17.5_{7.1}$ & $29.1_{4.4}$ \\
\hline
\multirow{5}{*}{tweet\_eval\_feminist} & 0\% & $48.7_{10.1}$ & $48.7_{10.1}$ & $56.3_{6.0}$ & $56.3_{6.0}$ & $55.6_{4.2}$ & $55.6_{4.2}$ & $25.6_{8.5}$ & $28.7_{7.2}$ \\
 & 25\% & $39.6_{10.8}$ & $39.6_{10.8}$ & $41.8_{7.0}$ & $42.7_{7.8}$ & $62.1_{3.7}$ & $61.6_{3.1}$ & $25.5_{7.7}$ & $26.1_{7.0}$ \\
 & 50\% & $37.7_{6.5}$ & $41.2_{5.2}$ & $41.1_{16.4}$ & $43.9_{13.9}$ & $56.8_{3.7}$ & $57.7_{3.3}$ & $28.4_{7.7}$ & $23.8_{10.0}$ \\
 & 75\% & $43.5_{8.5}$ & $47.3_{6.0}$ & $43.4_{12.2}$ & $51.6_{8.9}$ & $56.7_{4.1}$ & $56.9_{2.9}$ & $27.7_{8.8}$ & $21.0_{6.8}$ \\
 & 100\% & $44.4_{5.6}$ & $33.8_{12.0}$ & $49.7_{9.5}$ & $29.1_{11.7}$ & $47.3_{6.1}$ & $33.9_{6.8}$ & $37.7_{7.5}$ & $16.9_{2.1}$ \\
\hline
\multirow{5}{*}{tweet\_eval\_hate} & 0\% & $52.3_{5.3}$ & $52.3_{5.3}$ & $67.5_{3.6}$ & $67.5_{3.6}$ & $61.3_{1.8}$ & $61.3_{1.8}$ & $40.4_{5.1}$ & $40.3_{5.2}$ \\
 & 25\% & $51.0_{4.5}$ & $50.9_{4.5}$ & $66.5_{4.5}$ & $60.7_{12.4}$ & $61.1_{4.0}$ & $61.5_{3.9}$ & $46.2_{6.5}$ & $43.8_{6.5}$ \\
 & 50\% & $51.8_{8.5}$ & $49.9_{7.5}$ & $66.7_{4.2}$ & $52.0_{8.5}$ & $62.4_{2.3}$ & $64.8_{1.6}$ & $46.6_{6.7}$ & $34.2_{4.5}$ \\
 & 75\% & $44.7_{5.2}$ & $53.5_{7.7}$ & $54.2_{7.5}$ & $51.6_{5.2}$ & $55.7_{10.1}$ & $62.4_{1.6}$ & $42.6_{11.5}$ & $47.0_{10.5}$ \\
 & 100\% & $51.0_{11.6}$ & $37.1_{6.6}$ & $52.7_{7.9}$ & $35.9_{2.8}$ & $51.1_{4.8}$ & $40.0_{4.5}$ & $43.3_{9.7}$ & $36.1_{5.5}$ \\
\hline
\multirow{5}{*}{tweet\_eval\_irony} & 0\% & $57.0_{1.9}$ & $57.0_{1.9}$ & $58.9_{3.2}$ & $58.9_{3.2}$ & $63.6_{2.6}$ & $63.6_{2.6}$ & $49.6_{2.6}$ & $49.6_{2.6}$ \\
 & 25\% & $51.0_{7.3}$ & $50.9_{7.2}$ & $56.4_{8.6}$ & $55.5_{10.2}$ & $63.0_{3.8}$ & $63.2_{3.4}$ & $51.1_{3.1}$ & $52.2_{2.9}$ \\
 & 50\% & $48.6_{6.8}$ & $43.5_{8.1}$ & $51.2_{12.4}$ & $43.0_{10.7}$ & $62.6_{3.6}$ & $65.3_{2.4}$ & $47.9_{8.2}$ & $46.4_{7.8}$ \\
 & 75\% & $46.7_{7.2}$ & $40.3_{5.5}$ & $53.8_{9.8}$ & $43.9_{8.3}$ & $59.3_{6.3}$ & $55.3_{8.4}$ & $47.2_{5.1}$ & $43.6_{4.7}$ \\
 & 100\% & $43.7_{8.1}$ & $32.3_{1.3}$ & $51.6_{8.0}$ & $35.3_{2.4}$ & $52.1_{9.3}$ & $34.7_{0.7}$ & $44.5_{6.4}$ & $43.0_{6.4}$ \\
\hline
\multirow{5}{*}{tweet\_eval\_offensive} & 0\% & $59.4_{4.8}$ & $59.4_{4.8}$ & $64.5_{1.4}$ & $64.5_{1.4}$ & $65.0_{1.8}$ & $65.1_{1.8}$ & $53.2_{5.3}$ & $53.2_{5.4}$ \\
 & 25\% & $62.7_{2.0}$ & $62.7_{2.0}$ & $60.0_{4.5}$ & $59.9_{4.5}$ & $66.2_{0.9}$ & $66.4_{1.1}$ & $60.3_{5.9}$ & $62.7_{3.0}$ \\
 & 50\% & $60.2_{4.9}$ & $58.0_{4.8}$ & $62.8_{2.2}$ & $61.5_{3.2}$ & $65.7_{1.9}$ & $67.0_{1.7}$ & $60.4_{6.8}$ & $61.6_{4.3}$ \\
 & 75\% & $55.0_{13.8}$ & $39.4_{5.7}$ & $53.5_{12.0}$ & $48.7_{5.8}$ & $56.5_{13.9}$ & $63.2_{3.7}$ & $56.6_{14.4}$ & $56.9_{7.7}$ \\
 & 100\% & $54.4_{9.1}$ & $35.9_{2.6}$ & $61.1_{2.9}$ & $32.9_{2.7}$ & $60.5_{3.7}$ & $36.4_{4.3}$ & $60.9_{6.6}$ & $31.5_{3.4}$ \\
\hline
\end{tabular}
\caption{Results (Macro-F1) across datasets, models, and corrected proportions for ICL and CICL.}
\label{tab:detailed_results}
\end{table*}

\end{document}